# Topic Modeling Analysis of Aviation Accident Reports: A Comparative Study between LDA and NMF Models

Aziida Nanyonga
School of Engineering and Information Technology
University of New South Wales
Canberra, Australia
a.nanyonga@adfa.edu.au

Hassan Wasswa
School of Engineering and Information Technology
University of New South Wales
Canberra, Australia
h.wasswa@adfa.edu.au

Graham Wild
School of Engineering and Information Technology
University of New South Wales
Canberra, Australia
g.wild@adfa.edu.au

*Abstract*— **Aviation safety is paramount in the modern world, with a continuous commitment to reducing accidents and improving safety standards. Central to this endeavor is the analysis of aviation accident reports, rich textual resources that hold insights into the causes and contributing factors behind aviation mishaps. This paper compares two prominent topic modeling techniques, Latent Dirichlet Allocation (LDA) and Non-negative Matrix Factorization (NMF), in the context of aviation accident report analysis. The study leverages the National Transportation Safety Board (NTSB) Dataset with the primary objective of automating and streamlining the process of identifying latent themes and patterns within accident reports. The Coherence Value (C_v) metric was used to evaluate the quality of generated topics. LDA demonstrates higher topic coherence, indicating stronger semantic relevance among words within topics. At the same time, NMF excelled in producing distinct and granular topics, enabling a more focused analysis of specific aspects of aviation accidents.**

## I. Introduction

The aviation industry is a cornerstone of modern society, facilitating global travel, commerce, and cultural exchange. This sector continually advances its technology and safety protocols in pursuit of safer and more efficient operations ([1], [2]). However, despite these efforts, aviation accidents continue to occur, prompting rigorous investigations to discern the root causes and develop preventive strategies. These investigations yield voluminous textual data in the form of accident reports, which harbour invaluable insights into enhancing aviation safety. Extracting meaningful information from these reports, given their volume and complexity, poses a considerable challenge. In this context, topic modelling emerges as a potential tool for analyzing aviation accident reports. Topic modelling, a natural language processing (NLP) technique, unveils latent thematic structures within a corpus of text documents [3], [4]. By systematically categorizing and exploring the content of accident reports, it contributes to a deeper understanding of the factors underlying aviation accidents.

The motivation behind studying aviation accident reports is grounded in the aviation industry's unwavering commitment to safety enhancement. This commitment has led to significant advancements in technology, training programs, and regulatory measures. However, accidents persist, often resulting from intricate interplays between human factors, technical failures, and environmental conditions. Timely and precise analysis of accident reports is imperative for preventing future incidents and elevating aviation safety measures.

This paper aims to demonstrate the effectiveness of topic modeling, particularly through the application of Latent Dirichlet Allocation (LDA) and Non-negative Matrix Factorization (NMF) models, in the analysis of aviation accident reports. Our objectives are threefold:

1. To identify and extract latent topics present in aviation accident reports, providing a structured and interpretable representation of their content thereby shedding light on potential areas for safety improvement.

2. To conduct a comparative analysis of LDA and NMF models, assessing their performance in uncovering meaningful topics from accident reports.

This study is of great significance as it has the potential to enhance aviation safety through data-driven insights. By systematically analyzing a substantial corpus of accident reports, we aim to contribute to identifying recurring themes and contributing factors, informing regulatory enhancements, guiding training programs, and driving innovation in aviation technology. Our research leverages the NTSB Aviation Dataset, encompassing comprehensive accident reports spanning years between 2000-2020, and employs rigorous methodologies for data preprocessing, model training, and evaluation based on topic coherence and interpretability metrics.

In the following sections of this paper, we will delve into the related work, and our methodology, present the experimental results, and engage in a comprehensive discussion of our findings. Furthermore, we will explore the implications of our research for the field of aviation safety and suggest potential avenues for future research.

## II. RELATED WORK

First, In the realm of aviation safety analysis, the utilization of advanced techniques, such as topic modelling, is relatively novel but holds significant promise. Existing literature primarily focuses on traditional methods of accident investigation, expert analysis, and statistical approaches. However, a growing body of work acknowledges the potential of natural language processing (NLP) and machine learning techniques, including topic modelling, to glean insights from aviation accident reports [5], [6].

Historically, aviation accident investigations relied heavily on human expertise and manual examination of accident reports [7]. These methods, although valuable, are time-consuming and prone to human bias. Experts painstakingly sifted through textual narratives, findings, and recommendations to identify recurring patterns and contributing factors. While these approaches have yielded valuable insights, they are limited in their ability to handle the vast and complex aviation accident reports.

Recent developments in automated text analysis techniques have opened new avenues for aviation safety research. Researchers have recognized the potential of NLP and machine learning to extract actionable insights from textual data [4], [8]. Text mining and sentiment analysis have been applied to analyze reports such as online forums [9], [10]. These approaches offer the advantage of scalability and objectivity, reducing the human bias inherent in traditional analyses.

Topic modeling, a subset of NLP, has gained traction in aviation safety research. It offers a systematic way to uncover latent thematic structures within textual data, making it particularly suitable for aviation accident reports [11], [12]. [13] introduced Latent Dirichlet Allocation (LDA), a seminal topic modeling algorithm that has since become a cornerstone in this field. LDA has found applications in diverse textual datasets, ranging from news articles and social media content to scientific literature [14]–[16]. Its ability to unveil underlying topics and relationships has proven invaluable in numerous contexts. Also, [17] introduced Non-negative Matrix Factorization (NMF), a dimensionality reduction technique, that has found application in text mining and topic modeling. NMF has been employed in various studies to extract topics from text data, providing an alternative approach to LDA ([16], [18]).

A study by [12] employed LDA to extract topics from aviation safety reports and identified emerging safety concerns. Similarly, [6] explored topic modeling using LDA to categorize narratives in aviation accident reports, providing a structured representation of accident data. Also, [5] employed text-mining techniques to analyze aviation accident reports, focusing on identifying significant terms and phrases. Their study laid the foundation for applying computational methods to accident report analysis. The study in [19] introduced a framework that combined text mining and deep learning to automatically classify accident reports into categories based on accident prevention. Their approach showcased the potential for automating key aspects of accident analysis, offering efficiency and consistency.

In study [20] a compelling was presented on the application of structural topic modeling, specifically LDA, to aviation accident reports. The research demonstrated the feasibility of utilizing topic modeling to uncover latent themes within accident narratives. Moreover, it highlighted the potential for automating certain aspects of the analysis process, thereby increasing efficiency.

While individual studies have applied topic modeling to aviation safety, comparative studies between different topic modeling techniques, such as LDA and NMF, remain relatively scarce. Such comparative analyses offer valuable insights into the strengths and weaknesses of different approaches [17] The adoption of advanced techniques, including topic modeling, in aviation safety research is steadily growing. Researchers recognize the potential of these methods to provide deeper insights into the causes and contributing factors of aviation accidents. Our study builds upon this foundation by specifically evaluating the performance of LDA and NMF models in extracting meaningful topics from aviation accident reports using the NTSB dataset, adding to the growing body of knowledge in this area.

## III. METHODOLOGY

This section provides a detailed account of the methodology employed in this study, encompassing data collection, data preprocessing, the selection of topic modeling techniques (LDA and NMF), model training, and evaluation metrics.

### A. *Data Acquisition*

Aviation incident and accident investigation reports used in this study were exclusively sourced from the National Transport Safety Board (NTSB) dataset spanning years from 2000 to 2020. The dataset comprising a collection of more than 36,000 records in JSON format was obtained from the following source: https://www.ntsb.gov/Pages/AviationQuery.aspx. These reports encompass textual narratives, findings, and recommendations from NTSB investigations, making them an invaluable resource for the study.

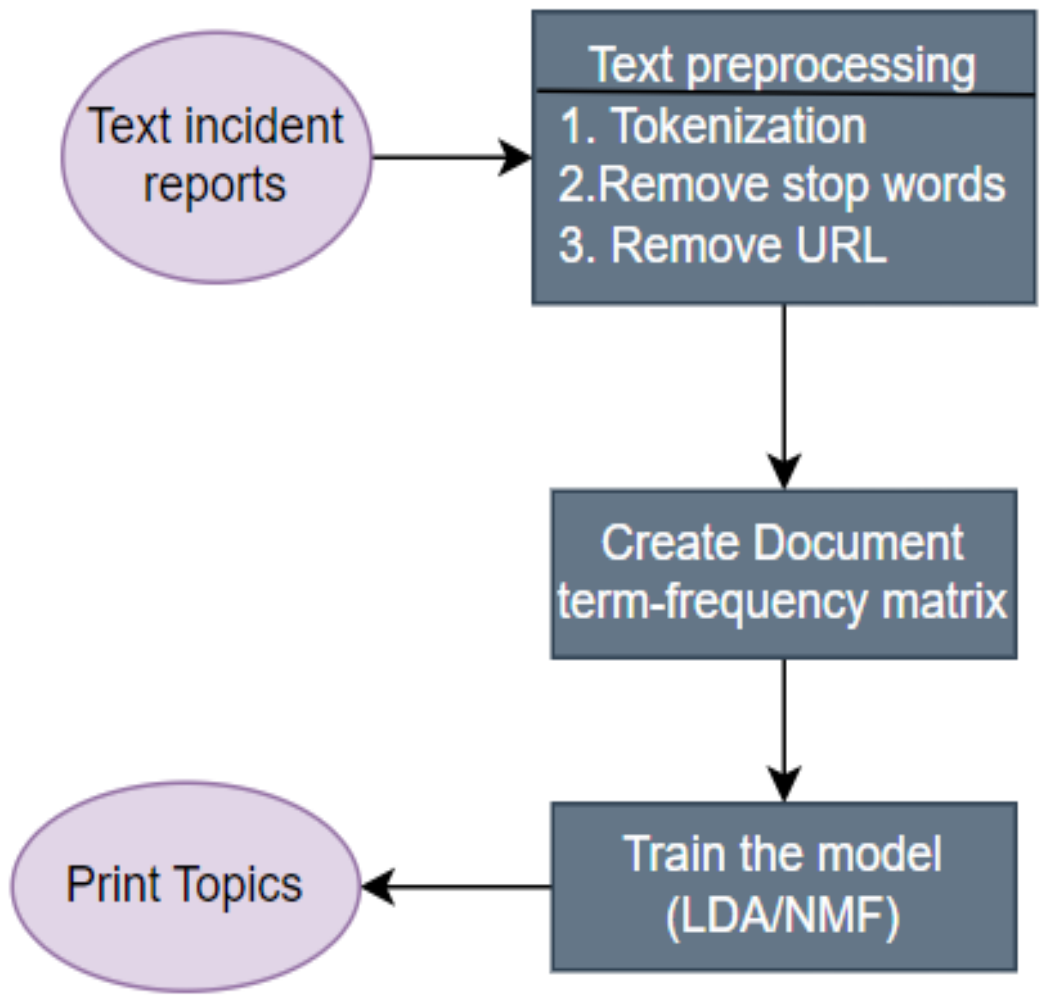

Fig. 1. Methodological framework

### B. *Text Processing*

Machine learning models inherently lack the capability to comprehend raw textual data. Our text preprocessing pipeline encompasses several essential stages to enhance data quality and improve model performance. These stages include tokenization, lowercasing, punctuation removal, stopword removal, and URL removal as shown in Fig. 1. Lowercasing ensures consistency in the text, while punctuation removal streamlines the text for analysis. Tokenization breaks down narratives into individual words, enabling further analysis. Stopword removal eliminates common stopwords, reducing noise, and URL removal ensures web links do not interfere with the analysis.

Once narratives undergo these preprocessing steps, they become ready for feature extraction a vital transformation that converts textual data into numerical features suitable for machine learning models. For feature extraction, we utilized two distinct techniques: Term Frequency-Inverse Document Frequency (TF-IDF) and Word Embeddings (Word2Vec). TF-IDF quantifies term importance within the narratives, capturing the semantic meaning of words. Word2Vec represents words as dense vectors, allowing models to understand semantic relationships. Additional preprocessing steps included the removal of HTML tags, non-alphanumeric characters, and other irrelevant elements. Finally, lemmatization reduced words to their base forms, improving topic modeling interpretability.

These comprehensive preprocessing and feature extraction steps ensure the textual data is transformed into a suitable format for subsequent topic modelling, leading to a more robust analysis of aviation accident reports.

### *C. Topic Modeling Procedure*

The preprocessed textual data were skillfully transformed into a Document-Term Frequency Matrix. This matrix plays a pivotal role in representing the frequency of each word across all narratives contained in the accident reports. Essentially, it provides a structured numerical representation of the textual data, facilitating the subsequent topic-modeling process.

#### *1) Latent Dirichlet Allocation (LDA)*

For the task of topic modelling, we harnessed the power of LDA, a probabilistic generative model that operates on the premise that documents are mixtures of topics, and topics, in turn, are mixtures of words. LDA is renowned for its established track record in topic modelling tasks [13]. This model excels at uncovering the latent thematic structures present within textual data, making it a natural choice for our analysis. Fig. 2 illustrates the operational mode of an LDA model. The model deploys a three-phase non-deterministic approach to assign topics to groups of words in each document. Phase one involves sampling a list of topics from a Dirichlet distribution of topics for each document [21] This is followed by phase two where each document word is assigned a topic from the sampled topics in phase one. Lastly, in phase three, each word assigned to a topic in phase two is sampled from a multinomial prior over words related to that topic. In this model, φ denotes the matrix of topic distributions, with a multinomial distribution over N-word items for each of T topics being drawn independently from a symmetric Dirichlet(β) prior. θ is the matrix of document-specific mixture weights for these T topics, each being drawn independently from a symmetric Dirichlet(β) prior. For each word, z denotes the topic responsible for generating that word, drawn from the θ distribution for that document, and w is the word itself, drawn from the topic distribution φ corresponding to z. $N_d$ stands for the number of words in the document. D stands for the size of the document collection. Estimating φ and θ provides information about the topics in a collection and the weights of those topics in each document.

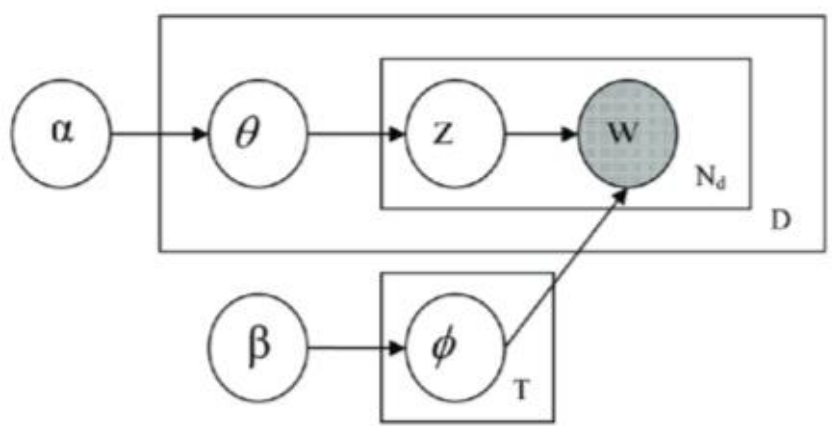

Fig. 2. Graphical representation of the LDA model [21]

#### *2) Non-negative Matrix Factorization (NMF)*

As an alternative dimensionality reduction technique in topic modeling, Non-negative Matrix Factorization (NMF) was also harnessed in our analysis. NMF factorizes the Document-Term Frequency Matrix into two lower-dimensional matrices one representing topics and the other representing term distributions [17]. What sets NMF apart is its innate interpretability, which is valuable in extracting meaningful insights from aviation accident reports. The dataset is represented as a $w \times d$, matrix, $V$. where $w$ represents words in each document, $d$. Fig. 3 illustrates a simple mechanism of how NMF breaks $V$ into its constituent components, $W$ and $H$ where $W$ is a $w \times t$ matrix, $H$ is a $t \times d$ matrix and t represents the distinct topics in $V$.

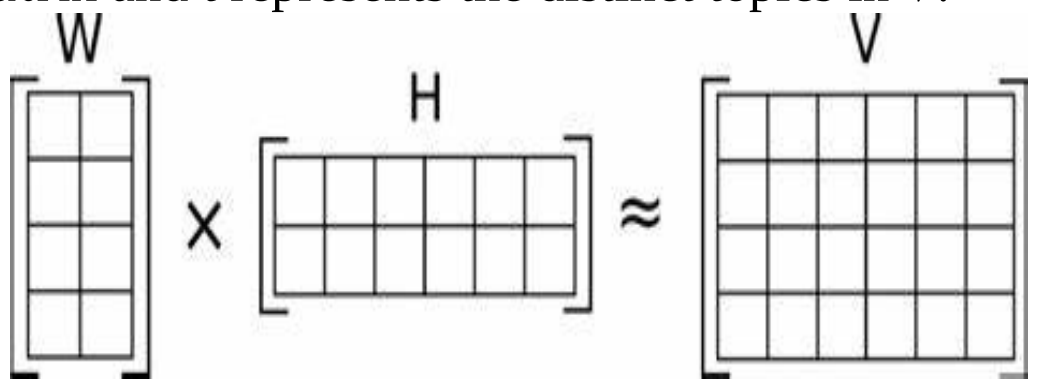

Fig. 3. Non-negative matrix factorization diagram

## IV. RESULTS AND DISCUSSION

In this study In this section, we present the results of our comparative study between LDA and NMF models in extracting topics from aviation accident reports using the NTSB Aviation Dataset.

### *A. Topic Extraction and Coherence Evaluation*

We employed the Coherence Value (C_v) as an evaluation metric to assess the quality of topics generated by the LDA and NMF models. The C_v measures the semantic coherence of topics, with higher values indicating more coherent topics. In our study, LDA yielded a C_v coherence score of 0.497, while NMF attained a C_v coherence score of 0.437.

#### *1) Topic Distribution*

We visualized the distribution of topics generated by both models. The topic distribution graph using the NMF model revealed distinct clusters of topics, suggesting meaningful separations. LDA also presented distinguishable topics, though with some overlapping themes as shown in Fig. 5 and Fig. 6, respectively. It is clearly seen that each of the ten topics was derived from a varying number of documents (narratives). This indicates that the models were able to precisely cluster the various documents, and ultimately assigned the appropriate topic to each document cluster. To facilitate a clear interpretation of the extracted information from a fitted LDA topic model, pyLDAvis was used to generate an intertropical distance map [18]. A screenshot of the statistical proximity of the topics can be seen in Fig. 4.

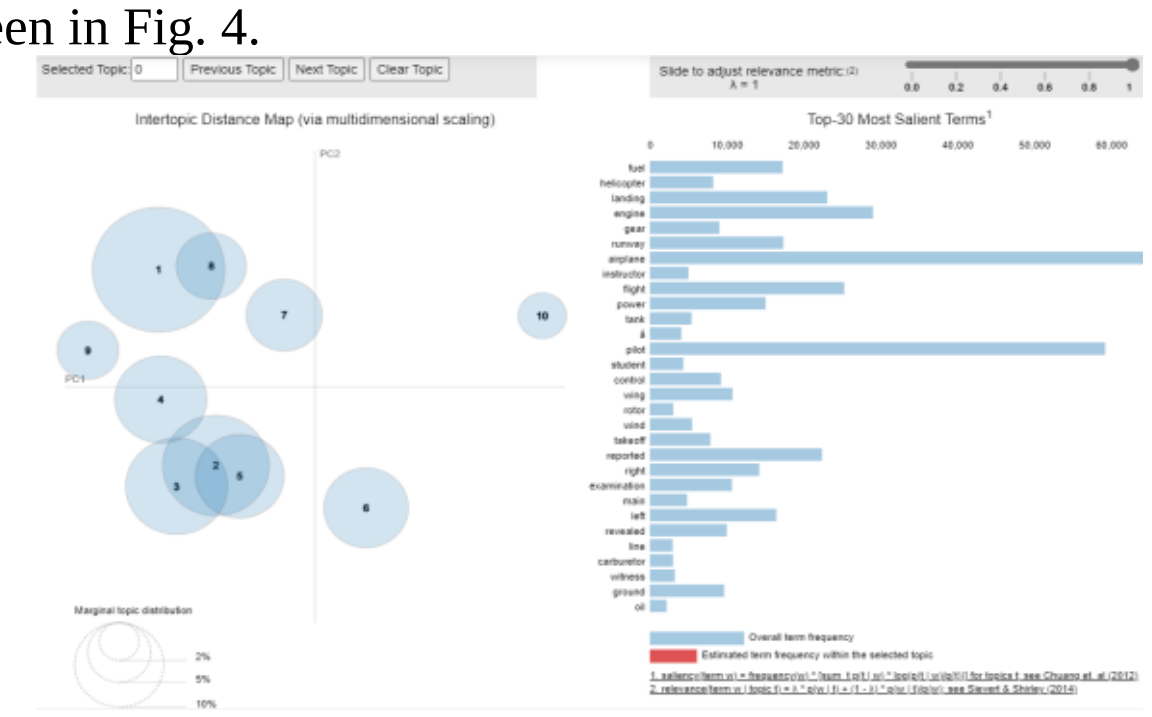

Fig. 4. Topic modeling visualization for aviation reports

TABLE I. TOP 5 WORDS FOR EACH OF THE 10 TOPICS THAT WERE CHOSEN BY BOTH MODELS

| Models | Topic 1 | Topic 2 | Topic 3 | Topic 4 | Topic 5 | Topic 6 | Topic 7 | Topic 8 | Topic 9 | Topic 10 |
|---|---|---|---|---|---|---|---|---|---|---|
| LDA | Pilot, | airplane | fuel | system | landing | engine | pilot | revealed | helicopter | airplane |
| | runway | pilot | tank | flight | gear | power | flight | examination | flight | pilot |
| | wind | reported | engine | control | pilot | airplane | airplane | inspection | instructor | accident |
| | knot | left | pilot | pilot | airplane | pilot | airport | bolt | pilot | flight |
| | accident | landing | airplane | seat | approach | loss | condition | fracture | rotor | witness |
| NMF | airplane | fuel | engine | helicopter | gear | flight | student | wind | right | runway |
| | reported | tank | power | rotor | landing | airplane | instructor | knot | left | airplane |
| | precluded | gallon | carburetor | tail | main | accident | flight | gust | airplane | Take off |
| | malfuncti on | engine | loss | blade | collapsed | condition | solo | gusting | rudder | approach |
| | operation | selector | icing | collective | nose | witness | cfi | accident | brake | foot |

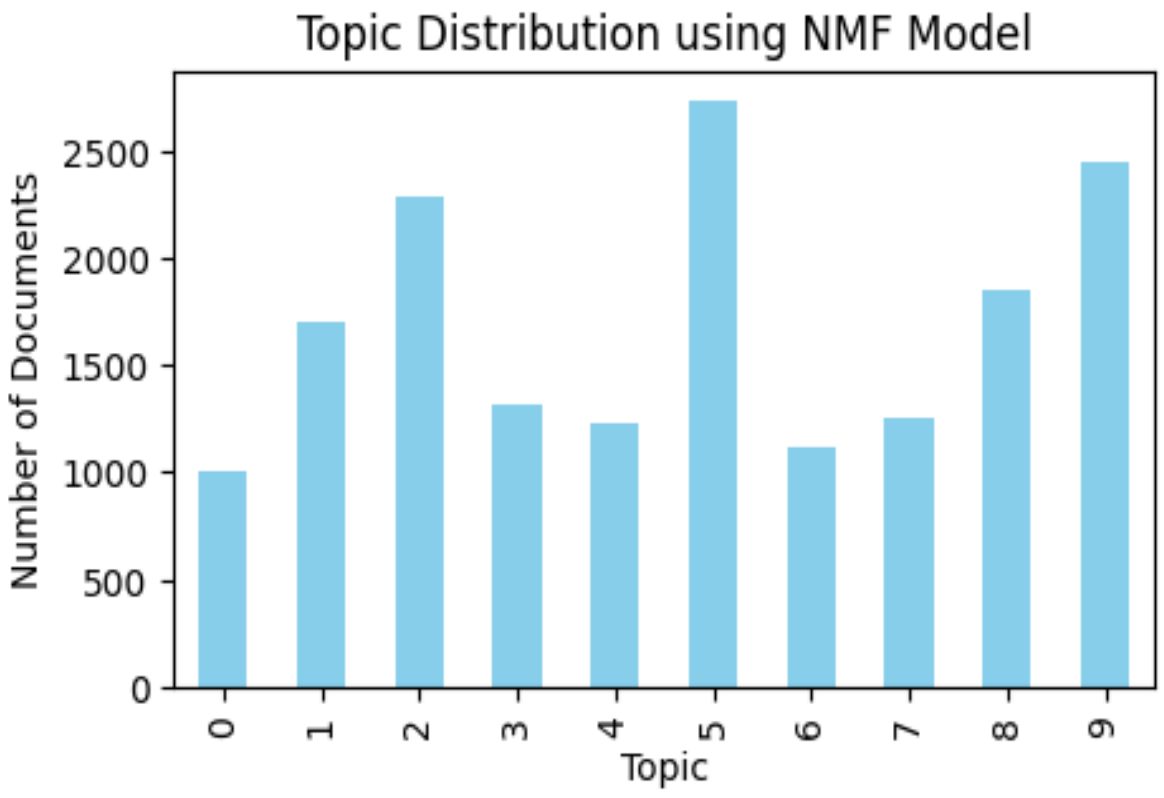

Fig. 5. Topic distribution on NMF Model

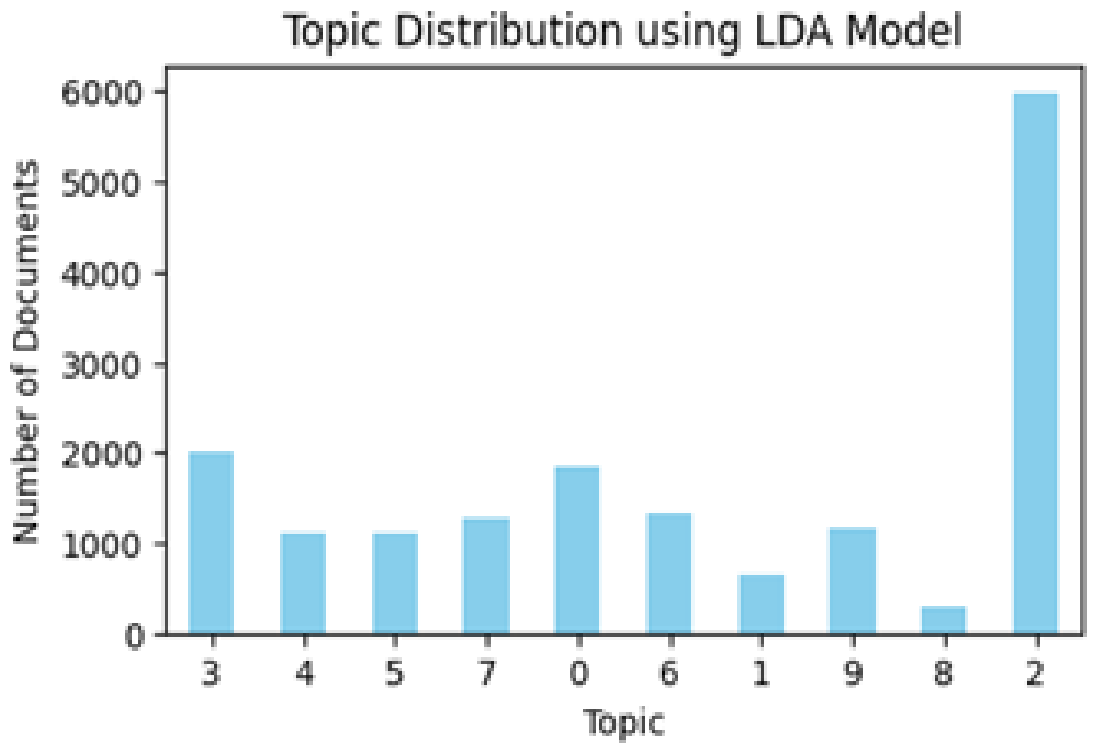

Fig. 6. Topic distribution on LDA Model

*2) Word Cloud and Top words*

To enhance topic interpretability, we provided word clouds for both LDA and NMF-generated topics. Table I shows the top 5 words for each of the 10 topics for each model. To facilitate understanding of the central themes captured by each model, the results in Table I are further visualized in Fig. 9 and Fig. 10 respectively. For each of the document clusters, the models learn the best defining topic words as shown in Fig. 11 for LDA and Fig. 12 for NMF. It should be noted that a topic word can appear in more than one topic. For instance, "pilot" appears in 8 out of 10 topics for LDA while "airplane" appears in 5 out of the 10 topics in NMF. This feature distinguishes topic modeling from conventional unsupervised learning-based clustering.

TABLE II. TOPIC EXTRACTION AND INTERPRETATION

| LDA-Generated Topics | NMF-Generated Topics |
|---|---|
| 1. Pilot and Aircraft Operations<br>2. Aircraft Mechanical Issues<br>3. Fuel Systems and Engine Performance<br>4. Flight Control Systems and Maintenance<br>5. Landing and Gear Operations<br>6. Engine Performance and Examination<br>7. Flight Conditions and Weather<br>8. Maintenance and Component Examination<br>9. Helicopter Operations and Instruction<br>10. Witness Reports and Accident Examination | 1. Aircraft Mechanical Failures<br>2. Fuel Systems and Engine Performance<br>3. Engine Performance and Examination<br>4. Helicopter Rotor and Tail Operations<br>5. Landing Gear Issues<br>6. Flight Conditions and Witness Reports<br>7. Flight Instruction and Solo Flights<br>8. Weather Conditions and Crosswind Effects<br>9. Aircraft Control and Runway Operations<br>10. Runway Operations and Takeoff |

For both models, the topics encompass a wide range of aviation-related themes, including aircraft operations, mechanical issues, fuel systems, engine performance, weather conditions, and flight instruction as shown in Table II. The topics extracted by LDA and NMF provide a comprehensive overview of the factors contributing to aviation accidents.

*3) Topic Co-occurrence Matrix*

We presented a topic co-occurrence matrix for both models, shedding light on the relationships and overlaps between topics as seen in Fig. 7 and Fig. 8.

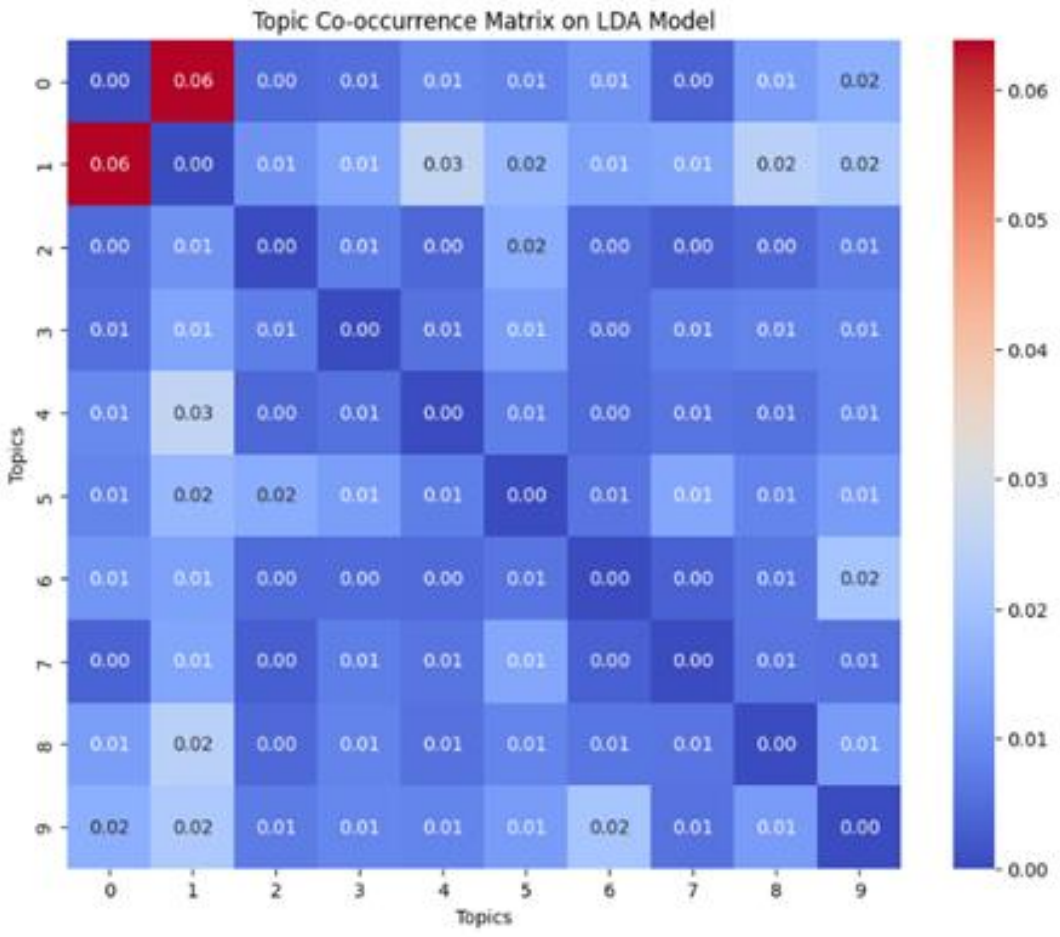

Fig. 7. Validation Accuracy for different models on both the largest and smallest NTSB dataset

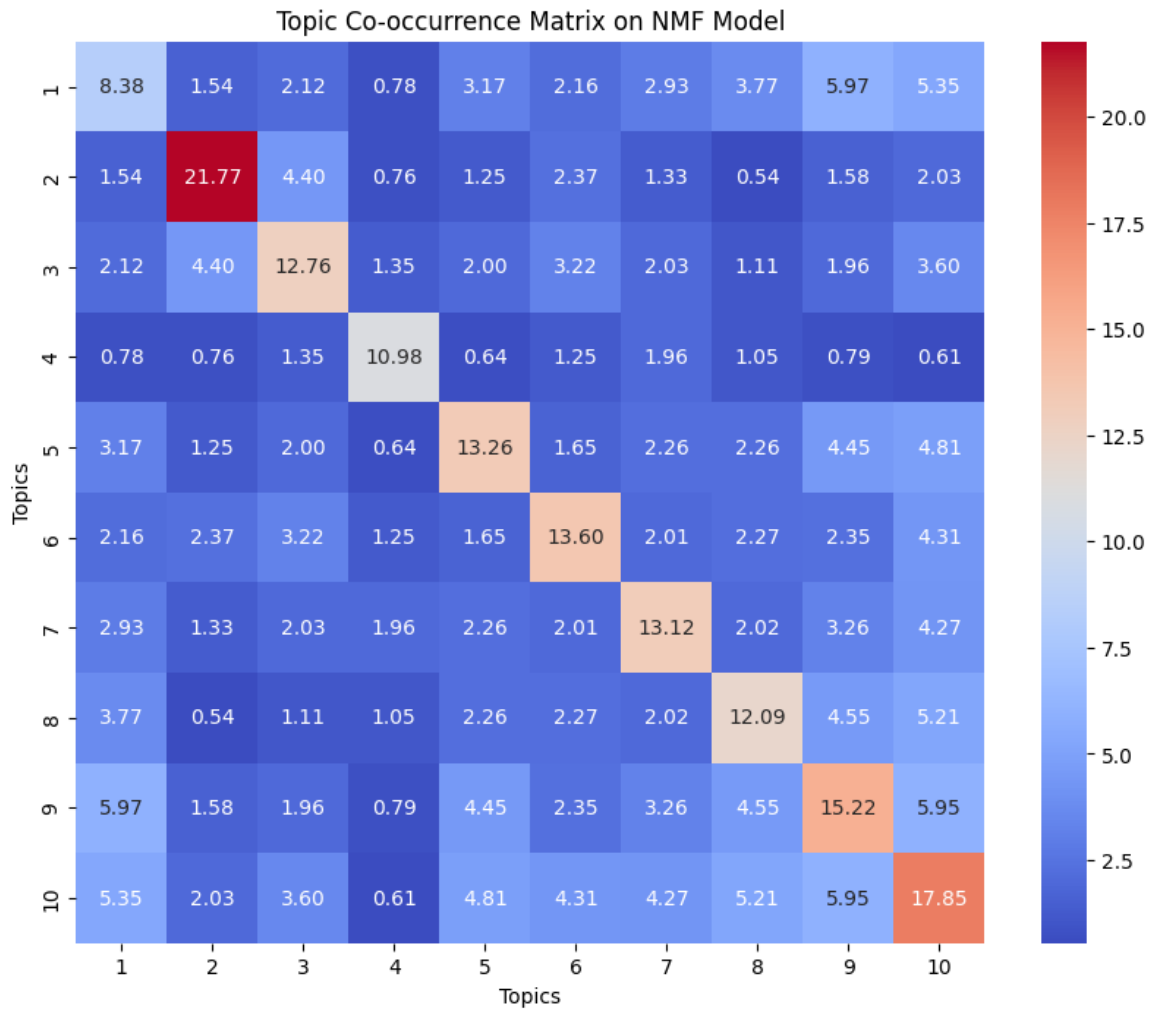

Fig. 8. Co-occurrence matrix

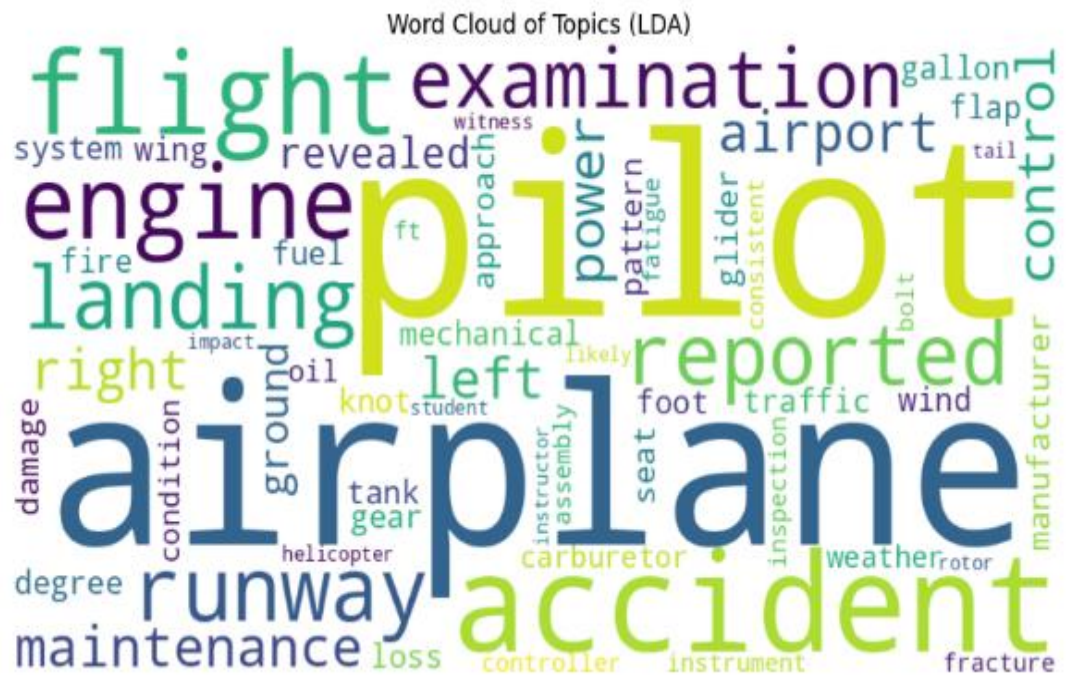

Fig. 9. LDA Topics

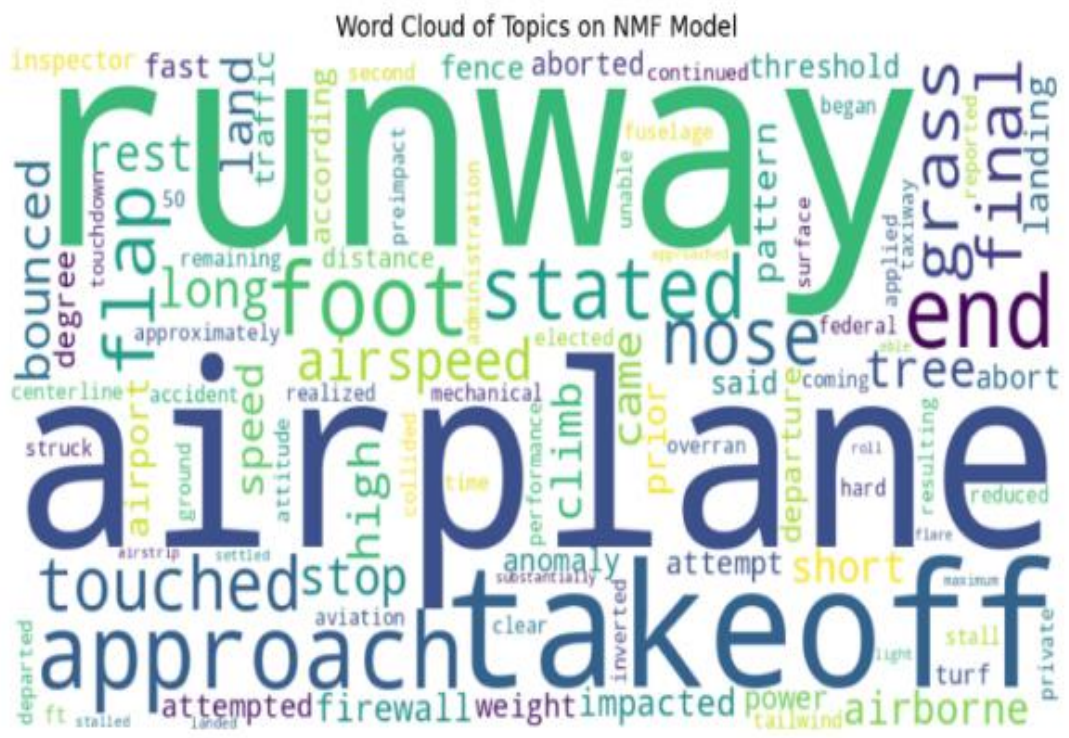

Fig. 10. NMF topics

Topic 1: airplane, reported, precluded, malfunction, operation, sustained, failure, normal, substantial, mechanical
Topic 2: fuel, tank, gallon, engine, selector, power, flight, pump, right, airplane
Topic 3: engine, power, carburetor, loss, icing, heat, oil, forced, examination, run
Topic 4: helicopter, rotor, tail, blade, collective, main, skid, ground, autorotation, rpm
Topic 5: gear, landing, main, collapsed, nose, extended, position, retracted, extend, revealed
Topic 6: flight, airplane, accident, condition, witness, instrument, airport, altitude, ft, turn
Topic 7: student, instructor, flight, solo, cfi, control, took, landing, instruction, instructional
Topic 8: wind, knot, gust, gusting, accident, weather, degree, crosswind, airport, runway
Topic 9: right, left, airplane, rudder, brake, wing, tailwheel, applied, veered, runway
Topic 10: runway, airplane, takeoff, approach, foot, end, stated, flap, nose, grass

Fig. 11. Shows the topics that were selected by NMF

Topic 1: pilot, runway, wind, knot, accident, airplane, reported, airport, foot, degree
Topic 2: airplane, pilot, reported, left, landing, runway, right, wing, damage, mechanical
Topic 3: fuel, tank, engine, pilot, airplane, power, flight, right, left, gallon
Topic 4: system, flight, control, pilot, seat, maintenance, fire, engine, accident, manufacturer
Topic 5: landing, gear, pilot, airplane, approach, runway, glider, traffic, pattern, flap
Topic 6: engine, power, airplane, pilot, loss, carburetor, examination, oil, revealed, landing
Topic 7: pilot, flight, airplane, airport, condition, accident, weather, instrument, ft, controller
Topic 8: revealed, examination, inspection, bolt, fracture, maintenance, assembly, fatigue, accident, consistent
Topic 9: helicopter, flight, instructor, pilot, rotor, student, reported, ground, control, tail
Topic 10: airplane, pilot, accident, flight, witness, likely, engine, ground, examination, impact

Fig. 12. Shows the topics that were selected by LDA

The comparative analysis reveals interesting findings regarding the performance of LDA and NMF in topic extraction from aviation accident reports. LDA outperforms NMF in terms of topic coherence, as indicated by the higher C_v score. This suggests that the topics generated by LDA exhibit stronger semantic relevance and coherence among the words within each topic. On the other hand, NMF exhibits its own strengths, particularly in selecting distinct and specific topics. NMF topics often focus on precise aspects of accidents, such as mechanical failures or weather conditions. This granularity can be advantageous for targeted analysis and decision-making. The word clouds figures, topic co-occurrence matrices figures, and top word table further illustrate the differences between the models' outputs. LDA tends to produce broader, more general topics, while NMF excels in capturing nuanced details.

## V. CONCLUSION

In this study, we conducted a comparative analysis of LDA and NMF models for topic modeling in the domain of aviation accident reports. We found that LDA achieved higher topic coherence, indicating a stronger semantic relationship between words within topics. However, NMF exhibited the ability to generate more specific and granular topics, which may be valuable for in-depth accident analysis. Our findings underscore the importance of considering the specific goals of topic modeling when choosing between LDA and NMF. Researchers and aviation safety experts can benefit from selecting the model that aligns with their objectives, whether it be comprehensive topic coverage or a detailed focus on specific aspects of accidents.

As aviation safety continues to evolve, the insights derived from topic modeling analysis can contribute to informed decision-making, enhanced safety measures, and a deeper understanding of the complex factors influencing aviation accidents. Future research may explore hybrid approaches that leverage the strengths of both LDA and NMF to further advance accident report analysis in the aviation industry.